\documentclass[conference]{IEEEtran}
\IEEEoverridecommandlockouts
\usepackage{cite}
\usepackage{amsmath,amssymb,amsfonts}
\usepackage{algorithmicx}
\usepackage{graphicx}
\usepackage{textcomp}
\usepackage{relsize}
\usepackage{multirow}
\usepackage{graphicx}
\usepackage{comment}
\usepackage{xcolor}
\def\BibTeX{{\rm B\kern-.05em{\sc i\kern-.025em b}\kern-.08em
    T\kern-.1667em\lower.7ex\hbox{E}\kern-.125emX}}

\begin{document}

\title{Pruning Literals for Highly Efficient \\ Explainability at Word Level\\
}

\author{\IEEEauthorblockN{Rohan Kumar Yadav}
\IEEEauthorblockA{\textit{Department of Information and Communication} \\
\textit{University of Agder}\\
Grimstad, Norway \\
errohanydv@gmail.com}
\and
\IEEEauthorblockN{Bimal Bhattarai}
\IEEEauthorblockA{\textit{Department of Information and Communication} \\
\textit{University of Agder}\\
Grimstad, Norway  \\
bimal.bhattarai@uia.no}
\and
\IEEEauthorblockN{Abhik Jana}
\IEEEauthorblockA{\textit{School of Electrical Sciences} \\
\textit{Indian Institute of Technology, Bhubaneswar}\\
Odisha, India \\
abhikjana@iitbbs.ac.in}
\and
\IEEEauthorblockN{Lei Jiao}
\IEEEauthorblockA{\textit{Department of Information and Communication} \\
\textit{University of Agder}\\
Grimstad, Norway  \\
lei.jiao@uia.no}
\and
\IEEEauthorblockN{Seid Muhie Yimam}
\IEEEauthorblockA{\textit{Department of Informatics} \\
\textit{Universität Hamburg}\\
Hamburg, Germany \\
seidymam@gmail.com}

}

\IEEEoverridecommandlockouts
\IEEEpubid{\makebox[\columnwidth]{979-8-3315-0498-4/24/\$31.00 ©2024 IEEE \hfill}
\hspace{\columnsep}\makebox[\columnwidth]{ }}
\maketitle
\IEEEpubidadjcol

\begin{abstract}
Designing an explainable model becomes crucial now for Natural Language Processing~(NLP) since most of the state-of-the-art machine learning models provide a limited explanation for the prediction. In the spectrum of an explainable model, Tsetlin Machine~(TM) is promising because of its capability of providing word-level explanation using proposition logic. However, concern rises over the elaborated combination of literals (propositional logic) in the clause that makes the model difficult for humans to comprehend, despite having a transparent learning process. In this paper, we design a post-hoc pruning of clauses that eliminate the randomly placed literals in the clause thereby making the model more efficiently interpretable than the vanilla TM. Experiments on the publicly available YELP-HAT Dataset demonstrate that the proposed pruned TM's attention map aligns more with the human attention map than the vanilla TM's attention map. In addition, the pairwise similarity measure also surpasses the attention map-based neural network models. In terms of accuracy, the proposed pruning method does not degrade the accuracy significantly but rather enhances the performance up to $4\%$ to $9\%$ in some test data.
\end{abstract}


\section{Introduction}
The majority of Natural Language Processing~(NLP) tasks are highly dependent on attention-based Deep Neural Networks (DNNs) models~\cite{DanilukRW017, Bahdanau2015NeuralMT, yang-etal-2016-hierarchical}. While attention mechanisms have been claimed to facilitate interpretability since their development, the question of whether this is true has just recently been a hot topic of discussion~\cite{Serrano2019IsAI, Wiegreffe2019AttentionIN}. In addition, it is recently suggested in \cite{jain-wallace-2019-attention}, \cite{Serrano2019IsAI}, and \cite{Wiegreffe2019AttentionIN} three separate ways for assessing the explainability of attention. Particularly, \cite{jain-wallace-2019-attention}'s study is based on the idea that explainable attention scores ought to be unique for each prediction while also being consistent with other measures of feature importance. Similarly, \cite{Serrano2019IsAI}~suggests that the relevance of inputs does not always equate to attention weights. While these studies raise important issues, they also use model-driven approaches to manipulate attention weights and then assess the post-hoc explainability of the generated machine attention.\par

One way to evaluate if the machine attention map~(MAM) correlates with the human attention map~(HAM) is to compare the similarity between these two \cite{sen-etal-2020-human}. However, due to the BlackBox nature that is lack of transparency in the model, the focus is now shifting to some interpretable models. One of the powerful rule-based interpretable models is Tsetlin Machine, which aims to reduce the gap between explainability and performance to a significant level \cite{yadav-etal-2021-enhancing}.

Tsetlin Machine (TM) has become an architecture of the choice for a vast range of NLP tasks especially in text classification such as Word Sense Disambiguation~(WSD) \cite{yadav2021wordsense}, Sentiment Analysis \cite{rohan2021AAAI}, Fake News Detection \cite{bhattarai2021explainable} and Document Classification \cite{7}. While TM has been a good alternative approach to traditional Deep Neural Networks~(DNNs) because of its transparent learning, the explainable rules that it offers tend to be huge in numbers making them difficult to comprehend. TM learns the pattern based on the combination of propositional logic called clauses and these collections of the clause make a pattern for a particular class~\cite{Granmo2018TheTM,Abeyrathna2021adaptivesparse,zhang2021convergence}. It has been arguably accepted that interpreting such a clause can give insight into the model for humans to understand the underlying concept of the task \cite{bhattarai2021word,rohan2022ijcai,blakely2021closed}. However, it has also been accepted that due to sparse Boolean bag-of-words (BOW) input representation, the model tends to learn negated literals in the majority making each clause very humongous and impractical to interpret.

\par Hence in this paper, we design a pruning technique to eliminate the unwanted literals\footnote{Literals are the form of input features either in the original or negated form such as a word ``good'' is an original form of the literal and ``$\neg$ good'' is the negated form that makes up a clause.} from the clause so that the clauses are more efficiently explained. The pruning method is named: pruning by the frequency of literals in the model. As a result of the pruning method, the propositional rules obtained from the set of clauses are shortened thereby making it easier for humans to comprehend. We obtain the Tsetlin attention map (TAM) from the pruned clause and evaluate it with HAM using the similarity measure. For evaluation of the model, we use the two popular metrics: Comprehensiveness and Sufficiency \cite{deyoung-etal-2020-eraser}. The proposed pruning method is not similar to the removal of stopwords that are predefined and are common in most cases. However, our approach statistically removes the unwanted form of the features to add explainability.

\par The main contributions of the paper are as follows:
\begin{itemize}
    \item We design a pruning method using the frequency of the literals in the clause in the model.
    \item We generate much shorter and more efficient clauses that are shorter, and compact which would lead to more explainable NLP models.
\end{itemize}

\begin{figure*}
\centerline{
\includegraphics[width=1\textwidth]{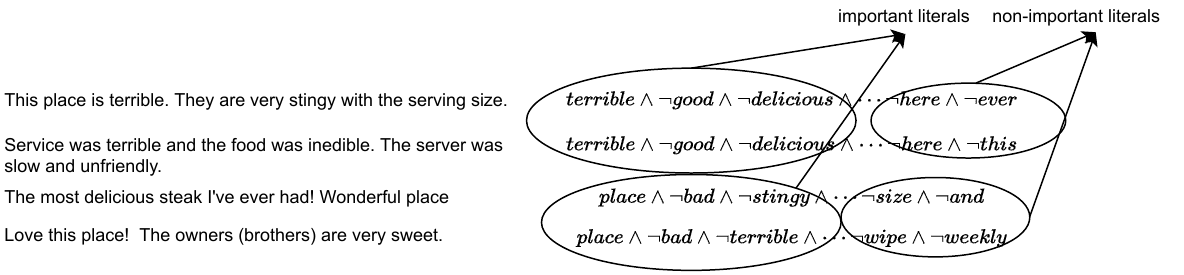}}
\caption{Example of important and non-important literals for given contexts.}
\label{fig1}
\end{figure*}

\section{Related Works}
We do not attempt to present a thorough overview of interpretability in NLP because it is a big and rapidly evolving field. Instead, we concentrate on directions that are particularly pertinent to our paper. Interpretability is a subjective aspect in the field of NLP. The important rationales for each individual may not agree with the other individual. This has been extensively studied in \cite{sen-etal-2020-human}. Thus interpretability has become a major evaluation metric for the NLP tasks apart from the accuracy of the model.

\par To extract rationales in contemporary neural networks for text classification, one might apply versions of attention \cite{Bahdanau2015NeuralMT}. Attention mechanisms learn to assign soft weights to (usually contextualized) token representations, and
so one can extract highly weighted tokens as rationales. Attention weights, on the other hand, may not always provide accurate explanations for predictions \cite{Serrano2019IsAI, Wiegreffe2019AttentionIN, jain-wallace-2019-attention}. This is most likely due to input entanglement in encoders, which makes interpreting attention weights on inputs over contextualized representations of the same more challenging.

\par Hard attention methods, on the other hand, discretely extract snippets from the input to feed to the classifier, resulting in accurate explanations. Hard attention processes have been offered as a technique for delivering explanations in recent research. \cite{lei-etal-2016-rationalizing} advocated creating two models, each with its own set of parameters, one to extract rationales and the other to consume them to make a prediction. \cite{jain-etal-2020-learning} developed a variant of this two-model configuration that employs heuristic feature scores to construct pseudo labels on tokens containing rationales; one model may then do hard extraction in this manner, while a second (independent) model can make predictions based on these.

\par Another area of research into interpretability is post-hoc explanation approaches, which aim to explain why a model generated a specific prediction for a given input. Typically, token-level significance scores are used. A common example is gradient-based explanations \cite{Sundararajan2017AxiomaticAF, Smilkov2017SmoothGradRN}. These have defined semantics that describes how locally perturbing inputs impact outputs. Since the local clause of TM deals with propositional logic, it does not have any mathematical soft or hard attention. Since the rules can be very large based on the number of vocabulary, it becomes a challenging task to understand the explainability it provides. Hence, in this paper, we propose a pruning method to get rid of the literals in the clause so that the model stills retains the valuable information but offers a much simpler explanation by getting rid of randomly placed literals.

\begin{figure}
\centerline{
\includegraphics[width=1\columnwidth]{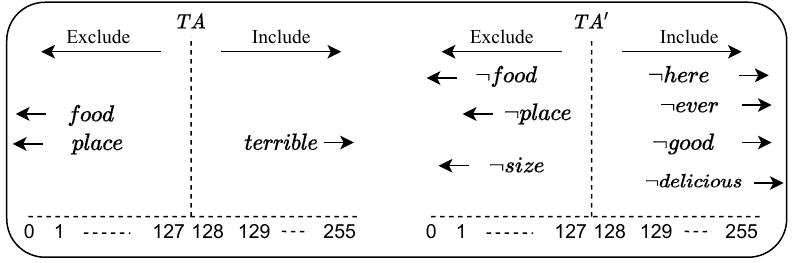}}
\caption{Representation of TA states for a given context.}
\label{fig2}
\end{figure}

\section{Proposed Methodology}
TM learns the patterns for a particular task by learning the sets of sub-pattern in the form of propositional logic known as a clause. During learning, TM has Tsetlin Automata (TA) states that decide which form of the literals take part in the clause. TM while learning includes some randomly initialized literals that do not impact the performance of the model. These randomly placed literals not being very important to either prediction class usually are in few numbers on count over the clause. Since their dominance is very weak in the model, we tend to reject it while evaluating the explainability of TM \cite{yadav2021wordsense}. We hypothesize that these randomly placed literals are non-important and do not impact the performance to a significant margin. If pruned from the clause, the TM still provides a similar accuracy. However, our main hypothesis is that since pruning reduces the size of propositional logic, the explainability can be easily comprehended by humans.

\subsection{Tsetlin Machine: Architecture}

The propositional logic in TM consists of the combination of the non-negated (original) and negated form of the literals in a conjunctive form. The selection of negated and non-negated forms of literals is decided by the set of TAs i.e., \textit{TA} controls the non-negated literals whereas \textit{TA'} controls the negated features. These actions of TAs are controlled by the two types of feedback namely: Type I Feedback and Type II Feedback \cite{rohan2021AAAI}. 
\par The most important aspect of TM \cite{Granmo2018TheTM} is the clause that learns the sub-pattern for each class using propositional logic. The sub-pattern is in the form of AND rules and is easily interpretable to humans. Here we will extract the interpretation of TM using the clauses of the model and generate a numerical attention map for each input feature. For instance, let us assume a Boolean bag-of-word \(X=[x_1, \cdots, x_n]\), $x_k\in\{0,1\}$, $k\in\{1,\ldots, n\}$ where $x_k=1$ means the presence of a word in the sentence and $n$ is the size of the vocabulary. If each class needs $\alpha$ clauses to learn the pattern and there are $\gamma$ number of classes, altogether the model is represented by \(\gamma \times \alpha\) clauses \(C_{\iota}^{\delta}\), \(1\leq {\delta} \leq \gamma\), \(1\leq {\iota} \leq \alpha\), as:
\vspace{-0.1cm}
\begin{equation}\label{eqn2}
    C_{\iota}^{\delta} = \left(\bigwedge \limits _{k \in I_{\iota}^{\delta}}{x_k} \right) \wedge \left(\bigwedge \limits _{k \in \bar I_{\iota}^{\delta}}{\neg x_k} \right),
\end{equation}
\noindent where \(I_{\iota}^{\delta}\) and \(\bar I_{\iota}^{\delta}\) are non-overlapping subsets of the input variable indices, \(I_{\delta}^{\iota}, \bar{I_{\delta}^{\iota}} \subseteq \{1, \cdots, n\}, I_{\delta}^{\iota} \cap \bar{I_{\delta}^{\iota}} = \emptyset\). These clauses vote for the respective class and keep the track of vote. Then a summation operator is used to aggregate the votes given by:

\vspace{-0.3cm}
\begin{equation}\label{eqn3}
    f^{\delta}(X) =\Sigma_{{\iota}=1,3,\ldots}^{{\alpha}-1}C^{\delta}_{\iota}{(X)}-\Sigma_{{\iota}=2,4,\ldots}^{\alpha}C^{\delta}_{\iota}{(X)}.
\end{equation}

\par For \(\gamma\) classes, the final output \(\hat y\) is given by the argmax operator to classify the input based on the highest net sum of votes, as shown in Eq~(\ref{eqn4}).
\begin{equation}\label{eqn4}
    \hat y =\mathrm{argmax}_{\delta}\left( f^{\delta}(X) \right).
\end{equation}

\subsection{Pruning Clauses in Tsetlin Machine}
TM uses a set of clauses to learn the collection of patterns in each class using the combination of literals which is detailed step-by-step in \cite{rohan2021AAAI}. Since TM is trained sample-wise, the literals that are included in the clause stay there for even longer training steps. This is because they do not have an impact on the prediction thereby not allowing the clause to be updated efficiently. An efficient clause is said to be the condition where a clause does not carry any randomly placed literals that do not have any impact on the prediction \footnote{Randomly initializing the TA states in TM is quite similar to initializing the weights randomly in neural networks.}.

\par \cite{yadav2021wordsense} has demonstrated that the literals that are less frequent in the clause over single or multiple experiments are said to be non-important literals. In contrast, the literals that are more frequent in the clause over the sample of single or multiple experiments are considered important literals. This means that there always exist some literals in the clause as a propositional logic that usually do not carry any significant information as shown in Fig.~\ref{fig1}.

\par Here in this paper we also validate the claim made by \cite{yadav2021wordsense}. Once the model is trained, we extract all the clauses along with the propositional logic associated with it. Note that: all the clauses from all the classes are considered. We understand that TA is the state that controls each form of the literal. For instance the states of TA for a context ``This place is terrible. They are very stingy with the serving size.'' are shown in Fig. \ref{fig2}. Here we can see that there are two TAs: \textit{TA} controls the states of the original form of the literals whereas \textit{TA'} controls the states of negated literals. Each TA state has two actions: Include and Exclude. When the states are from $0$ to $127$, the action Exclude is performed and the literals associated with corresponding TAs do not take part in the clause. On the other hand, the states are from $128$ to $255$ representing Include action where the corresponding literals now take part in the clause. From Figs. \ref{fig1} and \ref{fig2}, we can see that there are some literals in the negated form that are non-important. 

\par Our intuition is that the non-important literals such as $\neg here$ and $\neg ever$ should not impact the performance of the model significantly even if removed. However, important literals such as $\neg good$, $\neg delicious$ and $terrible$ should severely impact the model if removed. In order to categorize the important and non-important literals, we use the frequency of literals appearing in the set of clauses \cite{yadav2021wordsense} for a single experiment of selected epochs of training. We select the minimum $5\%$ to $50\%$ of the non-important literals and set their states to $0$ so that the desired literals are excluded from the calculation of prediction thereby pruning the clause. This study does not establish a benchmark for the number of literals that should be pruned but gives the study on how to explore pruning based on the required task.

\section{Experiments and Results}
Since our main aim is to prune the clause for better and significantly shorter propositional logic, we analyze the performance of the model based on two parameters: Primarily explainability and secondarily accuracy. There is various benchmark that provides human rationales for the evaluation of explainability \cite{Mathew2021HateXplainAB, sen-etal-2020-human}. However, we used the YELP HAT dataset \cite{sen-etal-2020-human}, a large-scale crowd-sourcing project that was devised and undertaken to collect human attention maps, which encode the elements of a text that humans focus on when undertaking text classification. It is collected for a specific classification task of the subset of the YELP dataset as positive or negative on Amazon Mechanical Turk. Three human attention maps have been collected for each sample of the YELP-HAT dataset given by HAM$_1$, HAM$_2$, and HAM$_3$. In addition to this, \cite{sen-etal-2020-human} provide a common training set of the data with three different subsets of testing data. The first subset consists of test samples with a length of 50 words named Yelp-50, the second subset of test samples with 100 words named Yelp-100, and the third subset of test samples with 200 words named Yelp-200. These subsets are designed because \cite{sen-etal-2020-human} established that the similarity between MAM and HAM changes based on the length of average words in the context.

\begin{table}
\centering
\caption{Pairwise Similarity Measure of comprehensiveness among HAMs and MAMs for Yelp-50. The proposed pruned TM consists of $\%$ of literals pruned.}
\label{tab:perf50}
\begin{tabular}{clclclclc}
\hline
\rule{0pt}{12pt}
&\textbf{Models} & \textbf{HAM$_1$} & \textbf{HAM$_2$} & \textbf{HAM$_3$}\\
\hline
\rule{0pt}{12pt}
    HAM &\textbf{HAM$_1$} &1 & 0.760 & 0.748 \\
   &  \textbf{HAM$_2$} & 0.76 & 1 & 0.765\\
   & \textbf{HAM$_3$} & 0.748 & 0.765 & 1 \\
   \hline
    \\[-6pt]
   NAM & \textbf{LSTM} & 0.621 & 0.643 & 0.634 \\
   & \textbf{Bi-LSTM} & 0.662 & 0.695 & 0.685 \\
&    \textbf{BERT} & 0.654 & 0.673 & 0.662 \\
\hline
\\[-6pt]
  TAM &  \textbf{vanilla TM} & 0.69 & 0.72 & 0.71 \\ 
    & \textbf{TM ($5\%$)} & 0.716 & 0.741 & 0.732 \\
    & \textbf{TM ($10\%$)}  & 0.718  & 0.743 & 0.739  \\ 
    & \textbf{TM ($15\%$)}  & 0.719 & 0.744 & 0.739  \\ 
    & \textbf{TM ($20\%$)}  & 0.721 & 0.765 & 0.743  \\ 
    & \textbf{TM ($25\%$)}  & 0.721 & 0.747 & 0.744  \\ 
    & \textbf{TM ($30\%$) } & 0.724  & 0.750 & 0.749 \\ 
    & \textbf{TM ($35\%$)}  & 0.722 & 0.748 & 0.747  \\ 
    & \textbf{TM ($40\%$)}  & 0.726  & 0.752 & 0.748  \\
\hline
\end{tabular}
\end{table}

\subsection{Explainability}

For the evaluation of explainability, we design an attention vector similar to a neural network. We address neural network-based attention maps as Neural Attention Map (NAM). MAM includes all the machine-generated attention maps including NAM and TAM in this case.  We obtain the TAM using comprehensiveness (were all features needed to make a prediction selected?) and sufficiency (do the extracted rationales
contain enough signal to come to a disposition?) \cite{deyoung-etal-2020-eraser}. We also calculate comprehensiveness and sufficiency as evaluation metrics for the neural network baselines i.e., LSTM, BiLSTM, and BERT. Here we use default parameters for BERT whereas we used Adam optimizer, dropout of $0.25$ with 256 hidden nodes for both LSTM and BiLSTM that achieved state-of-the-art performance for selected datasets.  To calculate comprehensiveness, we create a new input $\tilde{x}_k$ by removing the expected non-importance token $t_k$ from an original input $x_k$.  
Let $\mathcal{F}(x_k)_{\gamma}$ be the initial prediction by a model $\mathcal{F}$ for the predicted class $\gamma$. We then observe the prediction probability for the same class after $t_k$ is removed. Basically, when $t_k$ is important and removed from $x_k$, the model should be less confident in its prediction. This can be quantified as follows:
\vspace{-0.1cm}
\begin{equation}
\textit{comprehensiveness} = \mathcal{F}(x_k)_{\gamma} - \mathcal{F}(\tilde{x}_k)_{\gamma}.
\label{eq:comprehensiveness}
\end{equation}  

The score obtained from Eq.~(\ref{eq:comprehensiveness}) indicates the effect of $t_k$ in prediction. For example, a high score will indicate the removed tokens are of significant relevance, whereas a low score signifies the opposite. To calculate sufficiency, we measure if $t_k$ is sufficient for a model to make a prediction. The calculation is given as:

\begin{equation}
{\textit{sufficiency}} = \mathcal{F}(x_k)_{\gamma} - \mathcal{F}(t_k)_{\gamma}.
\label{eq:sufficiency}
\end{equation}  

In Eq.~(\ref{eq:sufficiency}), we assume the non-importance tokens $t_k$ for $\gamma$ class. Additionally, we adopt a simple approach to assess the faithfulness of scores assigned to tokens by a model. Eventually, we use both of this information to build an attention vector for the selected models. For DNNs baselines such as LSTM, Bi-LSTM, and BERT, we directly use the $\mathcal{F}$ as the probability of the model for the respective class. However, in the case of TM, we only have the clause score for each class. Hence we obtain the probability by normalizing the score obtained by subtracting the clause score of the sum of false classes from the clause score of a true class given by Eq.~(\ref{eqn1}).
\vspace{-0.1cm}
\begin{equation}\label{eqn1}
    \mathcal{F}(x_k) =  Norm\left(|f^{\delta=tp}(x_k) - \Sigma {f^{\delta=fp}(x_k})|\right).
\end{equation}

\noindent Here $f(x_k)$ represents the clause score of selected $(x_k)$ for true prediction \textit{tp} and false prediction \textit{fp}. |.| refers to the absolute value. Hence we map the TAM for comprehensiveness and sufficiency.

\begin{table}
\centering
\caption{Pairwise Similarity Measure of comprehensiveness among HAMs and MAMs for Yelp-100.  The proposed pruned TM consists of $\%$ of literals pruned.}
\label{tab:perf100}
\begin{tabular}{clclclclc}
\hline
\rule{0pt}{12pt}
 &\textbf{Models} & \textbf{HAM$_1$} & \textbf{HAM$_2$} & \textbf{HAM$_3$}\\
    \hline
    \\[-6pt]
    HAM & \textbf{HAM$_1$} &1 & 0.822 & 0.823 \\
    & \textbf{HAM$_2$} & 0.822 & 1 & 0.831\\
    & \textbf{HAM$_3$} & 0.823 & 0.831 & 1 \\
    \hline
    \\[-6pt]
    NAM & \textbf{LSTM} & 0.695 & 0.692 & 0.634 \\
    & \textbf{Bi-LSTM} & 0.714 & 0.713 & 0.710 \\
     &\textbf{BERT} & 0.766 & 0.764 & 0.762 \\ 
    \hline
    \\[-6pt]
    TAM & \textbf{vanilla TM} & 0.779 & 0.778 & 0.774 \\ 
    & \textbf{TM ($5\%$)} & 0.811 & 0.812 & 0.807 \\
    & \textbf{TM ($10\%$)}  & 0.819  & 0.820 & 0.815  \\ 
    & \textbf{TM ($15\%$)}  & 0.824 & 0.825 & 0.820  \\ 
    & \textbf{TM ($20\%$)}  & 0.828 & 0.828 & 0.822 \\ 
    & \textbf{TM ($25\%$)}  & 0.832 & 0.832 & 0.827  \\ 
     &\textbf{TM ($30\%$) } & 0.832  & 0.833 & 0.827 \\ 
    & \textbf{TM ($35\%$)}  & 0.833 & 0.833 & 0.828  \\ 
    & \textbf{TM ($40\%$)}  & 0.835  & 0.836 & 0.830  \\
    \hline
\end{tabular}

\end{table}

\subsubsection{Pairwise Similarity of Attention Map}
Here, we evaluate each respective attention with HAM using a pairwise similarity measure. For comparison, we use vanilla TM, LSTM, Bi-LSTM, and BERT as the baselines to compare with our proposed pruned TM. We selected these baselines because it covers the entire range from simpler to bigger model. From the collection of attention maps HAMs, MAMs (vanilla TAM and proposed pruned TAM and NAMs), the similarity measure between HAM and MAM is given by the average pair-wise similarity between each (HAM$_i$, MAM$_i$) where
\vspace{-0.1cm}
\begin{equation}\label{eqn:pair}
    PairSim_i = 1-|HAM_i -MAM_i|,
\end{equation}
\vspace{-0.5cm}
\begin{equation}
    SimMeasure = \frac{1}{|D|}\mathlarger\Sigma_i(PairSim_i).
\end{equation}

\noindent where $|D|$ is the number of reviews in the dataset D. This corresponds to using the human attention vector as binary ground truth, intuitively. In other words, it assesses how well the machine-generated continuous vector resembles the ground truth. The SimMeasure scores are between $0$ to $1$ with $1$ being perfect similarity and $0$ being no similarity. This measure gives the approximation of how far or close each sample of MAM is from gold HAM.

\textbf{Comprehensiveness}

\par We computed SimMeasure of comprehensiveness for proposed pruned TM and the selected baselines for Yelp-50 as shown in Table~\ref{tab:perf50}. It can be clearly seen that even HAMs differ from each other with a maximum SimMeasure of $.0765$. This shows that important rationales are subjective and may differ from human to human. When evaluating the baseline models LSTM, it aligns maximum with HAM$_2$ with a score of $0.643$  and least with HAM$_1$ with a score of $0.621$. A similar trend is seen in Bi-LSTM as well as BERT. The maximum SimMeasure between NAMs and HAMs is $0.695$. On the other hand, vanilla TM easily surpasses the SimMeasure of NAMs when compared to HAMs reaching a maximum of $0.72$. In addition to this, when the clauses are pruned with certain percentages starting from $5\%$ of the literals, the SimMeasure increases significantly thereby reaching $0.752$. The proposed pruned TM's similarity almost reaches that of HAMs.
\par Similarly, the SimMeasure of comprehensiveness for Yelp-100 is shown in Table \ref{tab:perf100}. Even in this case, vanilla TM's SimMeasure is better than that of NAMs. Interestingly, when the clauses are pruned more than $30\%$ of the literals, the SimMeasure outperforms the similarity among the HAMs as well. The same trend is observed in the case of the test sample having 200 words in the context as shown in Table \ref{tab:perf200}.

\begin{table}
\centering
\caption{Pairwise Similarity Measure of comprehensiveness among HAMs and MAMs for Yelp-200. The proposed pruned TM consists of $\%$ of literals pruned.}
\label{tab:perf200}
\begin{tabular}{clclclclc}
\hline
\rule{0pt}{12pt}
    &\textbf{Models} & \textbf{HAM$_1$} & \textbf{HAM$_2$} & \textbf{HAM$_3$}\\
    \hline
    \\[-6pt]
    HAM &\textbf{HAM$_1$} &1 & 0.831 & 0.858 \\
    &\textbf{HAM$_2$} & 0.831 & 1 & 0.854\\
    &\textbf{HAM$_3$} & 0.858 & 0.854 & 1 \\
    \hline
    \\[-6pt]
    NAM &\textbf{LSTM} & 0.665 & 0.652 & 0.684 \\
    &\textbf{Bi-LSTM} & 0.689 & 0.683 & 0.716 \\
    &\textbf{BERT} & 0.732 & 0.743 & 0.726 \\ 
    \hline
    \\[-6pt]
    TAM & \textbf{vanilla TM} & 0.841 & 0.829 & 0.875 \\ 
    & \textbf{TM ($5\%$)} & 0.823 & 0.811 & 0.856 \\
    & \textbf{TM ($10\%$)}  & 0.830  & 0.819 & 0.864  \\ 
    & \textbf{TM ($15\%$)}  & 0.844 & 0.831 & 0.877 \\ 
    & \textbf{TM ($20\%$)}  & 0.851 & 0.839 & 0.884 \\ 
    & \textbf{TM ($25\%$)}  & 0.845 & 0.833 & 0.878  \\ 
    & \textbf{TM ($30\%$) } & 0.851  & 0.839 & 0.885 \\ 
    & \textbf{TM ($35\%$)}  & 0.833 & 0.843 & 0.889  \\ 
    & \textbf{TM ($40\%$)}  & 0.835  & 0.845 & 0.891  \\
    \hline
\end{tabular}

\end{table}

\begin{table}
\centering

\caption{Pairwise Similarity Measure of sufficiency among HAMs and MAMs for Yelp-50. The proposed pruned TM consists of $\%$ of literals pruned.}
\label{tab:perf50gl}
\begin{tabular}{clclclclc}
\hline
\rule{0pt}{12pt}
&\textbf{Models} & \textbf{HAM$_1$} & \textbf{HAM$_2$} & \textbf{HAM$_3$}\\
\hline
\\[-6pt]
    HAM &\textbf{HAM$_1$} &1 & 0.760 & 0.748 \\
   &  \textbf{HAM$_2$} & 0.76 & 1 & 0.765\\
   & \textbf{HAM$_3$} & 0.748 & 0.765 & 1 \\
   \hline
    \\[-6pt]
    NAM &\textbf{LSTM} & 0.410 & 0.410 & 0.634 \\
    &\textbf{Bi-LSTM} & 0.537 & 0.536 & 0.534 \\
    &\textbf{BERT} & 0.527 & 0.534 & 0.529 \\ 
    \hline
    \\[-6pt]
    TAM &\textbf{vanilla TM} & 0.561 & 0.559 & 0.571 \\ 
    &\textbf{TM ($5\%$)} & 0.579  & 0.583 & 0.594 \\
    &\textbf{TM ($10\%$)}  & 0.585  & 0.582 & 0.599  \\ 
    &\textbf{TM ($15\%$)}  & 0.591 & 0.595 & 0.603 \\ 
    &\textbf{TM ($20\%$)}  & 0.598 & 0.615 & 0.620 \\ 
    &\textbf{TM ($25\%$)}  & 0.653 & 0.658 & 0.668  \\ 
    &\textbf{TM ($30\%$) } & 0.657  & 0.662 & 0.673 \\ 
    &\textbf{TM ($35\%$)}  & 0.657 & 0.672 & 0.677  \\ 
    &\textbf{TM ($40\%$)}  & 0.704  & 0.717 & 0.736  \\
    \hline
\end{tabular}

\end{table}

\textbf{Sufficiency}
For calculating the similarity measure we slightly change the Eq.~(\ref{eqn:pair}) slightly as given below:

\vspace{-0.2cm}
\begin{equation}\label{eqn:pairS}
    PairSim_i =|HAM_i -MAM_i|.
\end{equation}
This is because the lower the sufficiency, the higher the importance of the word in the model. We computed SimMeasure of sufficiency for the proposed pruned TM and the selected baselines for Yelp-50 as shown in Table \ref{tab:perf50gl}.  It can be seen that the SimMeasure of vanilla TM is better than the baseline NAMs. The increment is only marginal when the pruning is below $20\%$ but pruning more than $20\%$ of the literals significantly enhances the similarity thereby reaching $0.717$. A slightly interesting trend is observed for Yelp-100 as shown in Table \ref{tab:perf100gl}. Here the SimMeasure of vanilla TM is significantly lower than NAMs. However clause is pruned, and the SimMeasure enhances significantly. The highest SimMeasure for NAMs reaches $0.576$ by Bi-LSTM which is surpassed by pruned TM with more than $35\%$ of the literals. The same case is also observed for Yelp-200 where NAMs baselines are surpassed by pruned TM with more than $35\%$ as shown in Table \ref{tab:perf200gl}.

\begin{table}
\centering
\caption{Pairwise Similarity Measure of sufficiency among HAMs and MAMs for Yelp-100. The proposed pruned TM consists of $\%$ of literals pruned.}
\label{tab:perf100gl}
\begin{tabular}{clclclclc}
\hline
\rule{0pt}{12pt}
&\textbf{Models} & \textbf{HAM$_1$} & \textbf{HAM$_2$} & \textbf{HAM$_3$}\\
\hline
\\[-6pt]
    
    HAM & \textbf{HAM$_1$} &1 & 0.822 & 0.823 \\
    & \textbf{HAM$_2$} & 0.822 & 1 & 0.831\\
    & \textbf{HAM$_3$} & 0.823 & 0.831 & 1 \\
    \hline
    \\[-6pt]
    NAM &\textbf{LSTM} & 0.543 & 0.536 & 0.538 \\
    &\textbf{Bi-LSTM} & 0.576 & 0.560 & 0.564 \\
    &\textbf{BERT} & 0.552 & 0.551 & 0.554 \\ 
    \hline
    \\[-6pt]
    TAM &\textbf{vanilla TM} & 0.475 & 0.459  & 0.462 \\ 
    &\textbf{TM ($5\%$)} & 0.501  & 0.490 & 0.497 \\
    &\textbf{TM ($10\%$)}  & 0.527  & 0.514 & 0.526  \\ 
    &\textbf{TM ($15\%$)}  & 0.574 & 0.565 & 0.567 \\ 
    &\textbf{TM ($20\%$)}  & 0.556 & 0.548 & 0.555 \\ 
    &\textbf{TM ($25\%$)}  & 0.567 & 0.560 & 0.564  \\ 
    &\textbf{TM ($30\%$) } & 0.574  & 0.572 & 0.574 \\ 
    &\textbf{TM ($35\%$)}  & 0.636 & 0.635 & 0.634  \\ 
    &\textbf{TM ($40\%$)}  & 0.681  & 0.676 & 0.675  \\
    \hline
\end{tabular}

\end{table}

\begin{table}
\centering
\caption{Pairwise Similarity Measure of sufficiency among HAMs and MAMs for Yelp-200. The proposed pruned TM consists of $\%$ of literals pruned.}
\label{tab:perf200gl}
\begin{tabular}{clclclclc}
\hline
\rule{0pt}{12pt}
&\textbf{Models} & \textbf{HAM$_1$} & \textbf{HAM$_2$} & \textbf{HAM$_3$}\\
\hline
\\[-6pt]
HAM & \textbf{HAM$_1$} &1 & 0.831 & 0.858 \\
&\textbf{HAM$_2$} & 0.831 & 1 & 0.854\\
&\textbf{HAM$_3$} & 0.858 & 0.854 & 1 \\
\hline
\\[-6pt]
NAM & \textbf{LSTM} & 0.521 & 0.518 & 0.519 \\
&\textbf{Bi-LSTM} & 0.567 & 0.550 & 0.562 \\
&\textbf{BERT} & 0.517 & 0.511 & 0.515 \\ 
\hline
\\[-6pt]
TAM & \textbf{vanilla TM} & 0.473 & 0.483  & 0.450 \\ 
&\textbf{TM ($5\%$)} & 0.479  & 0.484 & 0.464 \\
&\textbf{TM ($10\%$)}  & 0.492 & 0.489 & 0.480  \\ 
&\textbf{TM ($15\%$)}  & 0.548 & 0.552 & 0.537 \\ 
&\textbf{TM ($20\%$)}  & 0.531 & 0.530 & 0.525 \\ 
&\textbf{TM ($25\%$)}  & 0.554 & 0.553 & 0.549  \\ 
&\textbf{TM ($30\%$) } & 0.554  & 0.549 & 0.549 \\ 
&\textbf{TM ($35\%$)}  & 0.609 & 0.609 & 0.605  \\ 
&\textbf{TM ($40\%$)}  & 0.606  & 0.603 & 0.605  \\
\hline
\end{tabular}

\end{table}

\subsubsection{Performance based on Accuracy}
We also evaluate the proposed method in terms of the accuracy of the model along with its baselines. Even though the main motive of this paper is to evaluate the explainability obtained from the models, we still need to obtain reasonable accuracy with a good trade-off between explainability and accuracy. Hence we obtain the accuracy of the model with the same configuration for which we obtained the comprehensiveness and sufficiency. We train each model on the same training data and test on three given test data.

\par Table \ref{tab:acc} shows the performance of the proposed pruned TM along with the selected baselines for given three subsets of test data Yelp-50, Yelp-100, and Yelp-200. We can see that both LSTM and Bi-LSTM-based model achieves similar accuracy on given test samples. However, BERT outperforms all of them thereby reaching $96.88\%$, $96.56\%$, and $93.48\%$ in Yelp-50, Yelp-100, and Yelp-200 respectively. Since TM is more comparable to LSTM and Bi-LSTM because these models do not have world knowledge as BERT does, we will focus on the rest of the models for comparison in terms of accuracy. Here, vanilla TM outperforms both LSTM and Bi-LSTM with more than $2\%$ in Yelp-50. However, its performance significantly declines for Yelp-100 with around $3\%$ and Yelp-200 with around $7\%$. This shows that the performance of TM  declines with the increase in the length size of the samples.

\begin{table}
\centering
\caption{Comparison of accuracy for TAM along with the pruned models and NAM.}
\label{tab:acc}
\begin{tabular}{clclclclc}
\hline
\rule{0pt}{12pt}
&\textbf{Models} & \textbf{Yelp-50} & \textbf{Yelp-100} &   \textbf{Yelp-200} \\
\hline
\multirow{3}{*}{NAM} & \textbf{LSTM} & 91.20 & 90.12 & 89.25 \\
&\textbf{Bi-LSTM} & 91.66 & 90.43 & 91.10 \\
&\textbf{BERT} & 96.88 & 96.56 & 93.48 \\ 
\hline
\\[-6pt]
\multirow{9}{*}{TAM} & \textbf{vanilla TM} & 93.33 & 87.57  & 82.66 \\ 
&\textbf{TM ($5\%$)} & 92.00  & 89.66 & 84.53 \\
&\textbf{TM ($10\%$)}  & 90.66 & 87.01 & 86.74  \\ 
&\textbf{TM ($15\%$)}  & 89.27 & 88.38 & 87.29 \\ 
&\textbf{TM ($20\%$)}  & 88.33 & 89.56 & 89.60 \\ 
&\textbf{TM ($25\%$)}  & 86.00 & 87.66 & 90.74  \\ 
&\textbf{TM ($30\%$) } & 85.00  & 89.34 & 91.53 \\ 
&\textbf{TM ($35\%$)}  & 83.50 & 90.23 & 88.23  \\ 
&\textbf{TM ($40\%$)}  & 85.66  & 91.12 & 86.54  \\
\hline
\end{tabular}
\end{table}

\begin{figure*}
\centerline{
\includegraphics[width=0.85\textwidth]{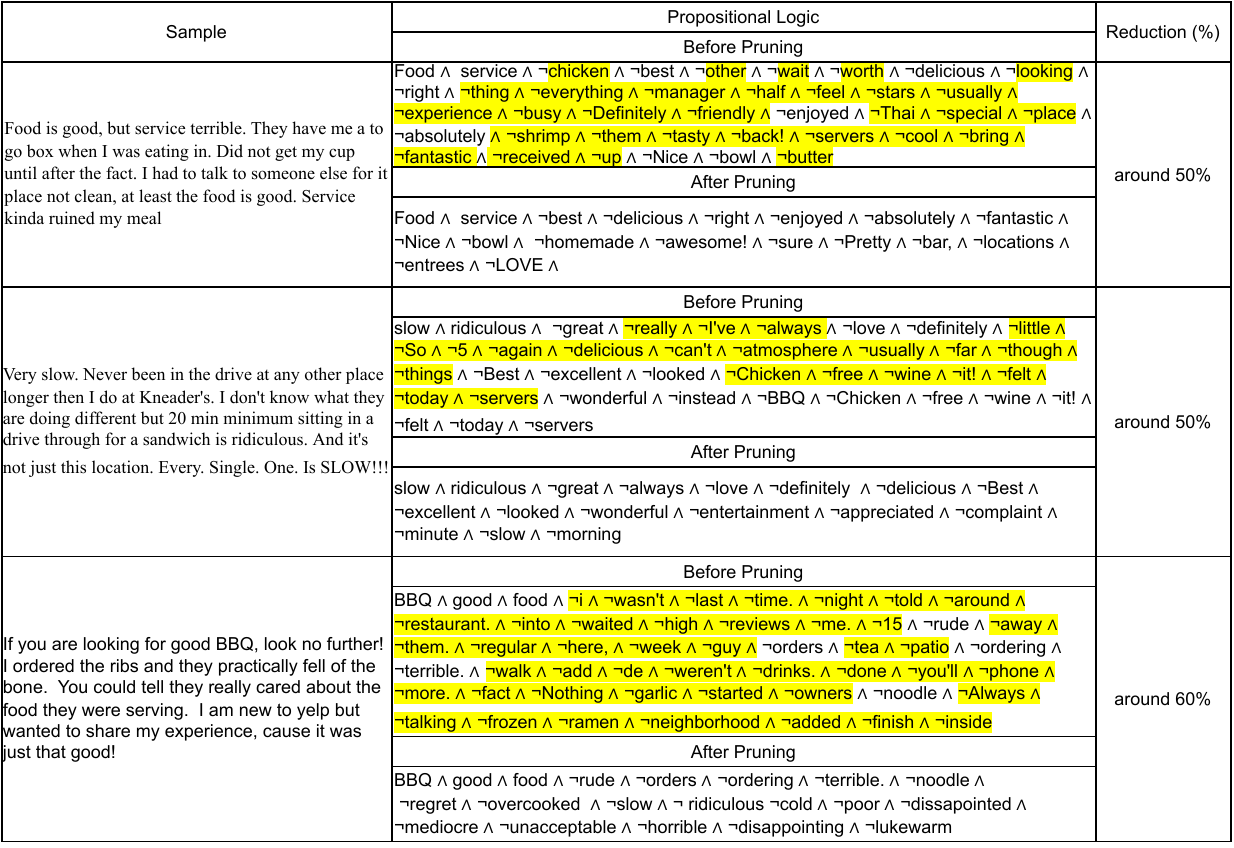}}
\caption{Case study for three samples and the corresponding clauses with propositional logic before and after pruning. Literals highlighters in yellow color represent that they are removed after the pruning.}
\label{fig3}
\end{figure*}

\par When we prune the trained clauses with certain percentages of non-important literals, the performance does not decrease drastically till the amount pruned literal is below $20\%$ in the case of Yelp-50. Since the decline in accuracy is very gradual, we can also see from Table \ref{tab:perf50} that the increment of SimMeasure is gradual as well. Similarly, for Yelp-100, the accuracy of vanilla TM is $87.57\%$
which is almost $3\%$ lower than the contender. However, when the clauses are pruned with $5\%$, the accuracy increase by $2\%$. This trend can be observed for the rest of the pruned TM as well. TM eventually reaches $91.12\%$ with $40\%$ non-important literals pruned thereby outperforming both LSTM and Bi-LSTM. This observation also supports the explainable part as we can see in both Table \ref{tab:perf100} and \ref{tab:perf100gl}, the SimMeasure drastically increases with high pruned literals. Similarly, in the case of Yelp-200, the accuracy of pruned TM achieves the highest of $91.53\%$ thereby outperforming vanilla TM by almost $9\%$ as well as outperforming LSTM and Bi-LSTM. This behavior is also observed in Table \ref{tab:perf200} and \ref{tab:perf200gl}, where the SimMeasure drastically increases for pruned TM thereby outperforming vanilla TM and other selected baselines.

\section{Impact on Explainable Propositional Logic}
In this section, we study the result of pruning in propositional logic. The main aim of pruning the clause is to obtain an easier and more compact propositional logic thereby retaining the accuracy of the model. Hence, we randomly select some samples and study their randomly selected clause. Each sample is passed to the trained model with and without pruning of the clause. This sample activates the clause to $1$ if satisfies the propositional logic.

\par Here we randomly select the clause of the predicted class for three selected samples are shown in Fig \ref{fig3}. The first two samples depict the negative sentiment, while the last sample represents the positive sentiment. Examination of the first sample reveals that the clause supporting the negative sentiment contains numerous non-important literals (as they do not carry the information about the sentiment of the context) such as $\neg other$, $\neg thing$, $\neg them$, $\neg wait$, $\neg looking$, etc.  Pruning $25\%$ of these non-important literals results in a more concise and highly explainable clause, featuring literals such as $\neg enjoyed$, $\neg homemade$, $\neg fantastic$, $\neg awesome$, etc., which are mostly sentiment-carrying words, leading to a $50\%$ reduction in overall literals. A similar outcome is observed for the second sample, where pruning resulted in a $50\%$ reduction of non-important literals like $\neg really$, $\neg I've$, $\neg always$, $\neg So$, etc. In the case of the positive sentiment sample, the removal of non-important literals such as $\neg i$, $\neg told$, $\neg times$, $\neg around$, etc. in a clause makes the pruned clause more compact with sentiment-carrying words in negated form, leading to a $60\%$ reduction in literals. This study highlights the presence of non-important literals that increase ambiguity in the explainability of the context. Removing these non-important literals provides a distinctive and concise proposition logic while maintaining accuracy and in some cases, improving it. In all brevity, if we use only the first few literals to explain the prediction of the model (such as 10-15 literals), we can observe that the clause before pruning has several literals that do not make sense to explain that something is negative in sentiment. Such literals are highlighted in yellow in Fig \ref{fig3}. Moreover, the pruning method not only removes non-important literals but also introduces new important literals such as ``$\neg$ disappointed'', ``$\neg$ mediocre'', ``$\neg$ horrible'', ``$\neg$ unacceptable'' etc as shown in the third sample having positive sentiment. The optimal proportion of literals to prune varies based on the task and dataset, but results suggest that pruning $20\%$ to $30\%$ of literals yields satisfactory results across all test datasets.
 
\section{Conclusion}

In this paper, we pruned the non-important literals that are randomly placed in the clause to enhance the explainability and performance of the model. We experimented on a publicly available dataset YELP-HAT that also consists of human rationales as the gold standard for evaluating explainability. The results demonstrated that pruning of clause significantly enhances the SimMeasure between TAM from pruned TM and HAM compared to TAM obtained from vanilla TM. Interestingly, pruned TM also enhances the accuracy from $4\%$ to $9\%$ in the case of Yelp-100 and Yelp-200 test data. In addition to this, we also demonstrated the impact of pruning on each randomly selected sample of both classes. We observed that pruning not only removes the unwanted literals but also makes compact and human-understandable propositional logic.

\section{Limitations}
Here, we have proposed a pruning of literals in TM that eliminated the non-important literals from the clause thereby making the proposition logic shorten and compact for better human explainability. However, there are still some concerns in TM that is a bottleneck for explainability. As we understand that TM can generate humongous propositional logic, even after shortening it by more than $50\%$, is still very huge in some cases. Such huge propositional logic is still difficult to comprehend for humans. Since, TM not only gives the weightage of each input feature on the model like DNNs, it incorporated high-level rules which can be subjective depending upon each individual. In addition to explainability, one of the limitations is the statistics of the dataset. Since TM does not employ world knowledge such as BERTs or GPTs, it might not categorize some words as important that do not occur very frequently in the dataset.

\nocite{*}
\bibliographystyle{IEEEtran}
\bibliography{References}
\end{document}